\documentclass[sigconf]{aamas} 
\renewcommand\footnotetextcopyrightpermission[1]{} 
\usepackage{balance} 
\usepackage{multirow}
\usepackage{float}
\usepackage{placeins}

\setcopyright{ifaamas}
\acmConference[AAMAS '25]{Proc.\@ of the 24th International Conference
on Autonomous Agents and Multiagent Systems (AAMAS 2025)}{May 19 -- 23, 2025}
{Detroit, Michigan, USA}{A.~El~Fallah~Seghrouchni, Y.~Vorobeychik, S.~Das, A.~Nowe (eds.)}
\copyrightyear{2025}
\acmYear{2025}
\acmDOI{}
\acmPrice{}
\acmISBN{}

\acmSubmissionID{<<649>>}

\title{Optimizing Large Language Models for Dynamic Constraints through Human-in-the-Loop Discriminators}

\author{Timothy Wei\footnotemark[1]\footnotetext{* These authors contributed equally to this work.}}
\affiliation{
  \institution{Saratoga High School}
  \city{Saratoga}
  \country{United States}}
\email{timswei@gmail.com}

\author{Annabelle Miin\footnotemark[1]}
\affiliation{
  \institution{Pacific Collegiate School}
  \city{Santa Cruz}
  \country{United States}}
\email{annabellemiin8@gmail.com}

\author{Anastasia Miin}
\affiliation{
   \institution{Pacific Collegiate School}
   \city{Santa Cruz}
   \country{United States}}
\email{anastasiamiin9@gmail.com}

\begin{abstract}
Large Language Models (LLMs) have recently demonstrated impressive capabilities across various real-world applications. However, due to the current text-in-text-out paradigm, it remains challenging for LLMs to handle dynamic and complex application constraints, let alone devise general solutions that meet predefined system goals. Current common practices like model finetuning and reflection-based reasoning often address these issues case-by-case, limiting their generalizability. To address this issue, we propose a flexible framework that enables LLMs to interact with system interfaces, summarize constraint concepts, and continually optimize performance metrics by collaborating with human experts. As a case in point, we initialized a travel planner agent by establishing constraints from evaluation interfaces. Then, we employed both LLM-based and human discriminators to identify critical cases and continuously improve agent performance until the desired outcomes were achieved. After just one iteration, our framework achieved a 7.78\% pass rate with the human discriminator (a 40.2\% improvement over baseline) and a 6.11\% pass rate with the LLM-based discriminator. Given the adaptability of our proposal, we believe this framework can be applied to a wide range of constraint-based applications and lay a solid foundation for model finetuning with performance-sensitive data samples.
\end{abstract}

\ccsdesc[500]{Computing methodologies~Multi-agent planning}
\ccsdesc[500]{Computing methodologies~Planning with abstraction and generalization}

\keywords{Large Language Model, Constraint-based Problem Solving, Discriminator, Finetuning}

\newcommand{\BibTeX}{\rm B\kern-.05em{\sc i\kern-.025em b}\kern-.08em\TeX}

\begin{document}


\pagestyle{fancy}
\fancyhead{}


\maketitle 

\section{Introduction}

The emergent intelligence of large language models (LLMs) has inspired researchers to explore their applications across diverse fields. However, besides foundational reasoning capability, different domains also naturally bring about unique application constraints that require LLMs to figure out and fulfill. Before the era of LLMs, constraint programming ~\cite{rossi2008constraint} was often used to transform constraint satisfaction problems into combinatorial search problems within the solution space, such as in travel planning, optimal scheduling, and network routing optimization. The rise of LLMs has prompted researchers to reconsider whether these problems can be approached more effectively using LLMs’ generalized capacity. 

Recently, reasoning patterns and model fine-tuning technologies, such as Chain-of-Thought (CoT) ~\cite{wei2022chain}, Tree-of-Thought (ToT) ~\cite{yao2024tree}, Self-Play finetuning ~\cite{chen2024self} have raised from the horizon and achieved great performance gains. Those methods usually rely on data curation reflecting on the deliberate reasoning path in specific application areas. When it comes to complex application constraints, high-quality solutions often demand a large volume of data to cover enough data cases and the corresponding reasoning logic. This process fundamentally differs from the typical human cognition process: faced with unfamiliar problems, people first seek to capture the overview of the underlying application constraints, which are potential rules summarized from observations. Next, these learned rules will be further refined when exception cases arise. The essence of this cognitive process lies in distilling rules and identifying minimal cases for refinement rather than depending on the inefficient generation of large datasets for training or fine-tuning to reach the desired performance. This is the primary motivation behind our proposal: the constraint-learning process should emulate this concept-to-optimization (CTOp) approach, ensuring data efficiency throughout.

Without losing generalizability, we chose travel planning ~\cite{xie2024travelplanner}, one of the common constraint problems, as the first problem to validate our idea. As shown in Figure~\ref{fig:framework}, our proposal is formalized to a two-phase learning procedure: the initial phase extracts constraints from system interfaces similar to the phase of grasping the required application constraints concept. Next, the optimization phase incorporates human experts and LLMs agents to identify critical cases and enhance agent performance that mimics the exception-case learning of human beings. Specifically, we model a human discriminator to represent human experts and an LLM discriminator to represent LLM agents, which unifies as discriminator agent shown in Figure~\ref{fig:framework}. To simplify the preliminary experiment and better isolate critical data influencing agent performance, we keep all agents in the zero-shot setting and left model finetuning for future work.

Our planning agent outperforms the baseline proposed by ~\cite{xie2024travelplanner} after a single feedback iteration. Further analysis of the LLM-based discriminator to identify critical cases shows that it effectively avoids selecting cases with minimal performance improvement.

In conclusion, the contributions of our paper are as follows:
\begin{itemize}
\item We introduce the concept-to-optimization (CTOp) approach and implement it as a learning process using a multi-agent collaboration process: The planning agent establishes constraint concepts and continuously optimizes its learning by collaborating with human and LLM discriminator agents to identify critical data cases. 
\item We evaluate our approach on the travel planning dataset ~\cite{xie2024travelplanner} and demonstrate its effectiveness through preliminary results.
\item We conduct a rigorous performance analysis of the planning and discriminator agents introduced in our proposal, identifying performance gaps between human and LLMs discriminator agents and laying down a solid foundation for future research.
\end{itemize}

\section{Related work}
Compared to existing paradigms for solving constraint-based problems in computer science, leveraging the reasoning and planning capabilities of LLMs remains relatively under-explored. While LLMs offer the potential for understanding natural language (NL) problem descriptions, efficiently utilizing their reasoning capacity to interpret constraints and determine optimal solutions poses significant challenges. The key issue lies in how to represent constraints for LLMs effectively, guide them to reason within those constraints, and assess the correctness or fallacies in their reasoning process.

\paragraph{Zero-shot Reasoning} This line of the work primarily focuses on generating comprehensive reasoning paths based on constraints and refining the LLM's performance by leveraging the most informative samples. One typical work is that \cite{gundawar2024robust} used the satisfiability modulo theory (SMT) framework and up to 10 back prompts to iterate and correct mistakes in user prompts. The tools they used (plan/itinerary generator, parser, and back prompter) were all LLM-based. Hao et al. ~\cite{hao2024large} also used the SMT framework and achieved a 97\% completion rate. They used LLMs primarily to parse and process the prompt with a feedback mechanism, used non-LLM SMT solvers, and called multiple APIs (for CitySearch, FlightSearch, AttractionSearch, DistanceSearch, AccommodationSearch, and RestaurantSearch) to incorporate structured itinerary data. However, their curated dataset was limited to flights and attractions and omitted detailed information on transportation methods, restaurants, and accommodations ~\cite{hao2024large} included in the original work ~\cite{xie2024travelplanner}. Similarly, ~\cite{hao2024large} and ~\cite{gundawar2024robust} also follow similar designs. In contrast, our approach aims to minimize the data required in the reasoning and argumentation process by first distilling the core constraint concept and then focusing on critical data samples during the optimization phase. We emphasize the potential to achieve incrementally better results using prompt discriminators. Additionally, this framework can be integrated with other methodologies and may generalize to a broader range of constraint-based problems beyond travel planning.

\paragraph{Finetuning based on Reasoning Data} Finetuning has been widely applied to improve the reasoning capacity of LLMs and even vision language models (VLMs). \cite{zhang2024scaling} discussed that LLM-based finetuning could encourage zero-shot generalization to relevant tasks, and parameter-efficient tuning could perform much better than full-model tuning. From our perspective, fine-tuning a model based on the constraints that most frequently fail can undoubtedly enhance reasoning capabilities. However, identifying the critical data points based on performance metrics while minimizing resource expenditure remains an under-explored area. \cite{lin2024data} discussed data-efficient finetuning and scoring data for 1) influential samples and 2) low-cost data pruning. They also focus on eliminating low-value data and focusing on higher-value data. While we did not score the data, we applied a scoring method to the user prompt to focus on higher-value data that address the high failure rate constraints. \cite{chen2024automated} focuses on automated data curation, employing a confidence-based system to estimate low-quality data and either filter or correct it. This has intriguing implications for feedback mechanisms aimed at improving user prompts. In contrast, our approach centers on leveraging a human-in-the-loop mechanism to identify critical cases for dynamic constraints, enabling broader applicability beyond fixed, pre-designed auto-metrics.

\paragraph{Multi-agent Collaboration} Researchers nowadays also design multi-agent systems to improve the final reasoning performance by realizing the limitation of single-agent system design. The motivation behind this is that complex tasks often require knowledge and expertise across multiple domains. Efficient collaboration among a group of specialized agents can maximize overall information gain and help prevent errors resulting from the knowledge gaps of individual agents. rStar~\cite{qi2024mutual} stride forward for self-play~\cite{zhang2024survey} reasoning of LLMs by incorporating two small language models (SLMs) without finetuning or superior models with a larger size, effectively addressing diverse reasoning problems. archon~\cite{saad2024archon} leverage a diverse set of LLMs and inference-time techniques, creating a modular framework for selecting, combining, and stacking layers of inference-time techniques to construct optimized LLM systems for target benchmarks. As a general framework, this method also supports us in creating more flexible systems, like ~\cite{zhang2024webpilot}, achieving excellent performance on web arena benchmark~\cite{zhou2023webarena}. The concept of multi-agent collaboration inspired us to introduce a discriminator agent, which identifies critical cases based on feedback from human experts and LLM agents. This enables the initial planning agents to evolve incrementally and ultimately meet the required performance thresholds. To the best of our knowledge, no prior work has used discriminators to quantify and score user prompts for improving efficiency and reducing data requirements in finetuning LLM-based models.

\section{Methodology}

\begin{figure*}[htbp]
  \centering
  \includegraphics[width=\linewidth]{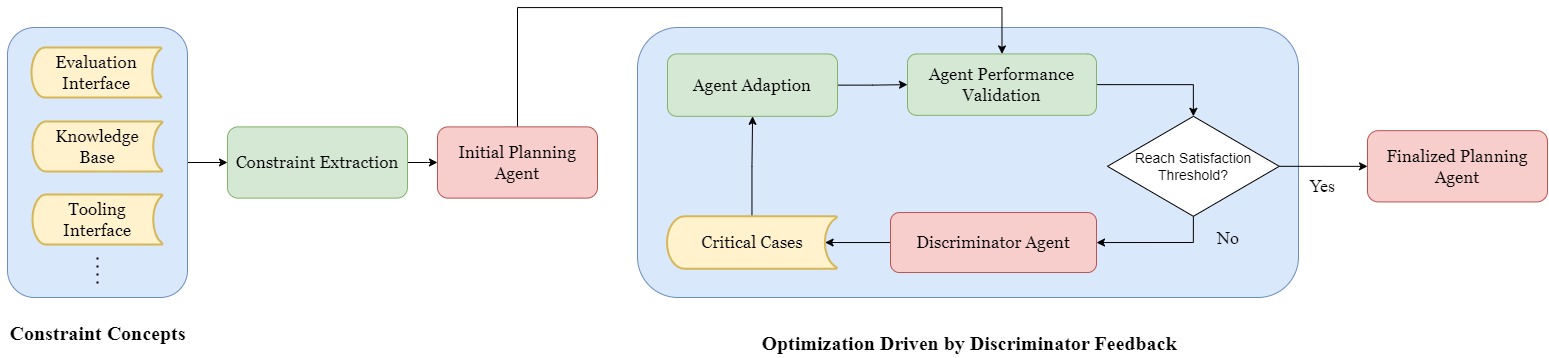}
    \caption{Our two-phase proposal for constraint concept building and continuous optimization driven by human expert annotation.}
    \label{fig:framework}
  \Description{Our two-phase proposal for constraint concept building and continuous optimization driven by human expert annotation.}
\end{figure*}

In this section, we present our method and demonstrate how it enables agents to evolve and meet performance requirements. As shown in Figure~\ref{fig:framework}, the planning agent is initially configured by extracting constraint concepts through interactions with system resources that reflect the underlying application constraints. The agent is then incrementally improved through a 'human-in-the-loop' iterative process, where critical cases are identified with the assistance of human experts and LLM discriminator agents. This iterative process continues until the planning agent meets the specified performance thresholds. The following sections detail the extraction of constraint concepts for initializing the planning agent and its optimization through interactions with prompt-based discriminators.

\subsection{Constraint Concept Building}
To enable LLM-based agents to comprehend application constraints and plan effectively, it is essential to extract and formalize these constraints in natural language (NL) format. We further categorize transformed constraints as either explicit or implicit, each arising from different aspects of the system.

\subsubsection{Explicit Constraints}
 Explicit constraints are usually rules or conditions defined in the system and must be satisfied. They can be directly obtained from the corresponding knowledge base, system application interface (API) specification, or other descriptive resources. They are usually clear, well-documented rules, such as resource limitations, scheduling requirements, or specific operational guidelines. In many cases, explicit constraints can be easily accessed through interfaces like evaluation tools or configuration systems, which provide formal descriptions to guide decision-making. These constraints are readily convertible into structured, textual formats, to seamlessly integrate into the planning agent during initialization.

\subsubsection{Implicit Constraints}
In contrast, implicit constraints are not immediately apparent and must often be inferred from the system's inputs and outputs through interactions with its API. Additionally, the transparency of the system's interfaces determines the difficulty of converting these constraints into the NL format. For simplicity, we consider two primary scenarios from the high-level perspective: 

\begin{enumerate}
    \item In a \textit{black-box system}, only the input-output (I/O) behavior is visible. Constraint extraction thus focuses on analyzing the patterns between the system’s inputs and outputs without knowledge of the internal mechanisms. For instance, based on recurring output patterns, like a refusal to schedule luxury hotels given an input budget of \$1000, we might deduce operational rules or limits that are not explicitly defined but govern the system's behavior, or in this case, that budget is a constraint considered the system. 
    \item In a \textit{white-box system}, the internal workings of the system, such as its source codes and workflow procedures, are fully accessible. This allows constraint extraction to go beyond just I/O interactions and consider intermediate steps, such as data processing or decision logic. However, the presence of intermediate computations introduces noise, making it harder to identify relevant constraints directly. 
\end{enumerate}

As explicit constraint resources have well-defined NL input and output, the corresponding concept building can be easily formulated as the few-shot learning or model finetuning for LLMs. Similar cases also fit for black-box systems because the pair of input and output are clear-cut and can processed in the same way. However, using LLMs to summarize and distill constraints from white-box systems involves challenges of how to avoid noises about intermediate processing, which is our main focus in this paper. Considering LLMs have been trained heavily via codes, such as~\cite{roziere2023code},~\cite{anil2023palm},~\cite{paik2021improving}, we suppose the State-of-The-Art(SOTA) LLMs are capable of understanding main functionalities of given code and extracting constraints from the given source code. We are also trying to leverage code analysis technologies, like abstract syntax tree analysis, dynamic compilation, and runtime interception, which provide more information about code control flow, variable dependencies, and runtime status to LLMs.
 
\subsection{Optimization Driven by Discriminator Feedback}
To further enhance the performance of the initial agent, we introduce a feedback loop involving two discriminator agents: a human discriminator agent and an LLM discriminator agent. Both discriminators review the data cases the initial agent processes and identify critical cases that can improve planning performance. Specifically, the discriminators rank data cases based on their assessed difficulty, allowing the planning agent to focus on these key cases to strengthen its ability to generate higher-quality plans.

Given the dynamic nature of system constraints, we do not rely solely on LLMs to evaluate data cases. Instead, we propose leveraging human expertise to define metrics for measuring the difficulty levels of data cases and to provide a reference ranking mechanism. Unlike LLM discriminators, which primarily rely on system feedback for ranking, human discriminators can better grasp the nuanced aspects of application constraints, helping the planning agent achieve significantly improved performance. Our preliminary results indicate that the human discriminator's contribution to performance improvements surpasses that of LLM discriminators. However, since human resources are often limited in practical systems, LLM discriminators offer distinct advantages in complex reasoning tasks, such as those seen in AlphaGo Zero~\cite{holcomb2018overview}. To balance their strengths, we introduce a hybrid model for discriminator agents, allowing them to jointly identify critical cases for enhancing the planning agent in subsequent iterations. The optimization loop continues until the planning agent meets the required performance threshold, at which point the agent reaches its finalized state.

\section{Experiments}

In this section, we outline the setup and describe the evaluation process, which consists of two stages: first, developing the initial agent, and second, enhancing the agent's performance through refined prompts provided by the discriminators.

\subsection{Stage I: Developing Initial Planning Agent}

The set-up at this stage included selecting an LLM to use to generate the constraints in step 1 of the diagram automatically and selecting an LLM to use for the travel-planning agent~\cite{xie2024travelplanner}. We decided to use GPT-4 Turbo as the base travel-planning agent model to make a fair comparison with the proposed agent in the travel-planning dataset ~\cite{xie2024travelplanner}. We chose to utilize GPT-4o to automate the generation of the constraints. Additionally, by utilizing the GPT models, we used the OpenAI API, which outsources the computing hardware required. 

 Specifically, we create an initial prompt for LLMs through its automatic summarization of Python code from the white-box system, which is the Python evaluation script in our case. This automatic step reduces labor intensity when creating the initial \textit{skeleton prompt} for use. This skeleton prompt will be used as the basis for the optimization module. We further use abstract syntax tree (AST) analysis to provide input and output pairs for methods in Python codes and feedback to LLMs for further improving the summarization result.

\subsection{Stage II: Improving Agent Through Refined Prompt Provided by Discriminators}
The prompt generated at stage I will be further refined by discriminators via selecting critical plans and adding to the prompt for improvement. This section discusses two discriminators, the human discriminator and the LLM discriminator, and how their performance is evaluated.

\subsubsection{Performance Evaluation of Discriminators}
To evaluate the performance of two discriminators, we used a set of randomly selected 10 plans to act as the discriminator data split. The task of the discriminators is to determine an ordering of those 10 plans that corresponded to how useful each plan would be for improving the performance if it was added as an example to the prompt in our framework. To obtain ground truth ordering, we ran each plan as an example into the validation data split 3 times and sorted the plans by the average Commonsense Constraint metric pass rate. 

\subsubsection{Human Discriminator}
This is a manual discriminator through human-in-the-loop. To create a human discriminator, plans were evaluated based on manual constraints listed in the query: number of days, number of cities, number of people, number of room rules (i.e. "must allow parties," "must be pet-friendly," etc.), number of cuisines (i.e. "American," "Indian," etc.), and transportation requests (i.e. "no self-driving," "no flights," etc.). Each constraint was then assigned a "difficulty score," with "0" being the easiest for an LLM to plan and any number "N," where N is greater than 0, having increased difficulty. These individual "difficulty scores" for each constraint were then summed into a "total difficulty score," with larger numbers correlating to plans with higher difficulty.

Difficulty scores for each constraint were assigned in the following way:
\begin{itemize}
  \item Number of days: Given that there are only three types of options: 3-day, 5-day, and 7-day, "0" was assigned to 3-day queries since they are the easiest types of queries a LLM could encounter, "1" was assigned to 5-day queries, and "2" was assigned to 7-day queries. 
  \item Number of cities: Since there are three options: 1-city, 2-city, and 3-city, "0" was assigned to 1-city queries since they are the easiest option a LLM could encounter, "1" was assigned to 2-city queries, and "2" was assigned to 3-city queries.
  \item Number of people: There could be a large number of people listed in the query, so we followed a similar methodology in assigning a difficulty score as the last two constraints: "0" was assigned to 1-person queries since they are the easiest types of queries a LLM could encounter, and for every plan that had more than 1 person, we assigned a difficulty score of "N-1," where "N" is the number of people listed in the query.
  \item Number of room rules: The difficulty score assigned equals the number of room rules since the easiest query possible has "0" room rules requested.
  \item Number of cuisines: The difficulty score assigned equals the number of cuisines since the easiest query possible has "0" cuisines requested.
  \item Transportation requests: The difficulty score assigned is either "0" or "1" since the easiest query has no special transportation requests and the hardest query only has 1 transportation request.
  \end{itemize}
Note that we assumed each constraint would have equal weighting in terms of difficulty.

The difficulty score is also used to rank which queries are most suitable for finetuning an LLM at a specific difficulty level. For example, if the goal is to finetune the algorithm to handle more complex queries, plans with higher difficulty scores should be prioritized for better results. Similarly, if the goal is to fine-tune the algorithm for easier queries, plans with lower difficulty scores should be utilized to optimize performance.

\subsubsection{LLM-based Discriminator}

In comparison, we also wanted to know how an automated discriminator agent, through a SOTA LLM, performs for screening out critical cases. To rank distinct plans based on their predicted pass rate, we inputted the plan and the evaluation code, which was obtained from the benchmark TravelPlanner~\cite{xie2024travelplanner}. The model we chose for this LLM discriminator here is gpt-4o-mini. The LLM was prompted to create parallels between certain plan parts and the evaluation code. For example, if there were a section of the code checking for cuisine constraints, then if the plan has a cuisine constraint, the LLM would rate this plan more highly than a plan without this cuisine constraint. The LLM discriminator then assigned each plan a number from 1-100, with 1 being the worst plan and 100 being the best plan to add as an example. We ran the discriminator on our 10 plans for each plan 10 times, computing an average score for the ultimate LLM-based discriminator score. To evaluate the LLM-based Discriminator, we use the metric "worst plan avoidance." This metric counts the number of times the discriminator avoided ranking the worst-performing plan as the top plan. In this case, we define the worst plan as contributing least to constraint knowledge when added to the prompt refinement. 

\subsection{Evaluation Implementation}

We utilize the evaluation method in TravelPlanner ~\cite{xie2024travelplanner} to evaluate the performance of our travel planning agent. This method evaluates the plan generated by the TravelPlanner using several constraints, split between \textit{\textbf{Commonsense Constraints}} and \textit{\textbf{Hard Constraints}}. Commonsense Constraints consist of those constraints that a traveler is not required to make, but will naturally make. For example, one commonsense constraint could be to require Diverse Attractions, as it would not make sense for multiple days to visit the same attraction. Hard Constraints, on the other hand, are things that human travelers are required to follow. These include constraints such as the budget for the trip as well as abiding by all the room rules of each accommodation that is booked. 

For baseline, we used the original TravelPlanner implementation. We call these results the \textit{\textbf{GPT-4 Turbo Baseline}}. Additionally, we used their zero-shot prompt, with unstructured reference data, as a baseline to compare to. We resorted to the data splits from the same paper, and these splits can be obtained in our GitHub. Specifically, to evaluate our results, we utilized the validation split with 180 queries, and to manually refine the automated prompt, we utilized the 45-query training split.

After analyzing the GPT-4 Turbo Baseline results, we manually identified and included poorly performing constraints in the prompt, leading to significant performance improvements. This resulted in the "gold standard" for our automation module to aspire to, which we call the \textit{\textbf{GPT-4 Turbo Manually Modified Prompt}}.

Previous papers' trials of the travel-planning agent utilized reference information in the form of raw JSON ~\cite{xie2024travelplanner}. While the JSON itself is structured, the information contained within each JSON key was actually in CSV form, so we felt that restructuring the reference information as a series of CSV files would allow the travel-planning agent to perform better. To this end, we created a script that converted each raw JSON reference information into a list of CSV files. Then, we appended this structured reference information to each prompt to achieve better results.

\section{Performance Evaluation}

\begin{table*}[htbp]
\centering
\caption{Agent Performance on Validation Dataset}
\label{main_result}
\setlength{\abovecaptionskip}{0.2cm}
\setlength\tabcolsep{3pt}
\resizebox{\textwidth}{!}{
\begin{tabular}{|l|c|cc|cc|c|} 
\toprule
    & \multirow{2}{*}{\textbf{Delivery Rate}} & \multicolumn{2}{c|}{\textbf{Commonsense Pass Rate~(\%)}} & \multicolumn{2}{c|}{\textbf{Hard Constraint Pass Rate~(\%)}} & \multirow{2}{*}{\textbf{Final Pass Rate}}  \\ \cline{3-6} 
    & & \textbf{Micro} & \textbf{Macro} & \textbf{Micro} & \textbf{Macro} & \\ 
\midrule
GPT-4 Turbo Baseline & 100.0 & 80.07 & 17.78 & 50.23 & 28.33 & 5.55 \\
GPT-4 Turbo Manually Modified Prompt & 100.0 & \textbf{84.93} & \textbf{27.22} & \textbf{61.19} & \textbf{42.78} & \textbf{12.78} \\
Initial GPT-4o Generated Prompt & 100.0 & 81.39 & 15.56 & 37.14 & 22.22 & 2.78 \\
Improved GPT-4o Prompt after LLM Discriminator & 100.0 & 67.92 & 16.67 & 14.29 & 13.33 & 6.11\\
Improved GPT-4o Prompt after Human Discriminator & 100.0 & 70.69 & 11.11 & 24.05 & 13.11 & 7.78 \\
\bottomrule
\end{tabular}}
\end{table*}

We collected agent performance statistics in Table \ref{main_result} and compared them with baselines. Table \ref{main_result} below shows the passing rates. Row "GPT-4 Turbo Baseline" shows the result from TravelPlanner ~\cite{xie2024travelplanner}. The next three rows contain the results from manually modified prompt, initial GPT-4o automated prompt and the improved prompt after 1st iteration with "human-in-the-loop".

From Table \ref{main_result}, we can see that the initial GPT-4o automatically generated prompt produces unsatisfactory result, lower than the GPT-4 Turbo Baseline. However, with the addition of the LLM discriminator, we see an improvement of the final pass rate from 5.55\% to 6.11\%.  Additionally, the Human Discriminator causes an even greater increase; from 5.55\% to 7.78\%. The way we evaluated these discriminators is we chose the highest ranking plan in each and added it to the skeleton prompt to determine results. The resulting performance metrics are now greater than the GPT-4 Turbo Baseline. In the meanwhile, the GPT-4 Manually Modified Prompt performs much better than the GPT-4 Turbo Baseline, achieving greater pass rates across the board as well as a final pass rate that is 2.3 times that of the baseline (12.78\% vs. 5.55\%). With one single iteration of "human-in-the-loop", we are getting closer to the performance of the GPT-4 Manually Modified Prompt and we are hopeful the gaps will become smaller with more iterations.

 \subsection{Discriminator Results}
 
 \begin{table*}[htbp]
 \caption{Human Discriminator Results}
 \label{r^2-results}
 \begin{tabular}{|l|cc|cc|c|}
 \hline
 \multirow{2}{*}{}   & \multicolumn{2}{c|}{\textbf{Commonsense ($R^2$)}} & \multicolumn{2}{c|}{\textbf{Hard Constraint ($R^2$)}} & \multirow{2}{*}{\textbf{Final Pass Rate ($R^2$)}} \\ \cline{2-5}
    & \textbf{Micro} & \textbf{Macro} & {\textbf{Micro}} & \textbf{Macro} & \\ 
     \midrule 
     Human Discriminator & 0.691 & 0.787 & 0.753 & 0.199 & 0.313 \\ \hline
 \end{tabular}
 \end{table*}

To understand the performance discrepancy between human and LLM discriminators, we compared discriminator performance statistics in Table \ref{r^2-results} below. While the LLM discriminator did not perform very well, the LLM Discriminator avoided ranking the worst plan as the best plan in each case that we tested it on. Meanwhile, human discriminators have reasonably good performance. Table \ref{r^2-results} shows the result of the Human Discriminator in ranking the plans. It has a $R^2$ of greater than 0.5 on all metrics. Thus, we conclude that for data selection for prompt-tuning, human discriminator quality is still the top tier. However, it's important to remember that the LLM-discriminator we have is very general, as it only took the evaluation code to make decisions, meaning that this technique can be applicable on a wide variety of domains.

\subsection{Key Findings}
Our analysis revealed several key findings regarding agent performance. Firstly, manual refinement of the initial prompt demonstrably enhanced the performance of the agent. This suggests that careful curation of the prompt plays a crucial role in guiding the agent towards generating desirable outputs. Secondly, both human experts and LLM discriminators proved effective in identifying useful plans for prompt refinement. This indicates the potential for human intuition and automated techniques to contribute to the prompt engineering process. Finally, restructuring the reference information from JSON to CSV format resulted in notable performance improvements. This highlights the importance of data formatting in facilitating efficient information processing by the agent.

\section{Limitations}
Our framework relies mostly on automation to generate prompts and evaluate the results. It also uses discriminator feedback to improve agent performance. However, the limitation and potential risks of our work also can't be neglected: 

\textbf{Limited data} Since the reference data in ~\cite{xie2024travelplanner} is limited in selecting travel information, accommodations, restaurants and attractions. Despite its generality, it currently cannot consider unsafe or incorrect information. We didn't have sufficient time to include a more diverse set of reference data for the agent. One potential risk is that it could generate imperfect plans based on incorrect information from the database.

\textbf{Prompt design} We need a careful design of instruction steps and corresponding code to design the prompt. It is time-consuming to formulate from scratch and we don't fully explored automatic prompt generation technology, such as Dspy~\cite{khattabdspy}, TextGrad~\cite{houtextgrad}. Also, we used our evaluation script to assess the efficacy of the resultant LLM-generated travel plans; however, LLMs don't inherently have this capability built in. 

\textbf{Cross domain application} Although we have demonstrated the applicability of this framework in travel planning, we have not yet tested it in other domains with distinct features and constraints, such as planning a high school or college student’s multi-year course load while accounting for budgetary, degree requirements, and other limitations. These untested cases may require more advanced capabilities beyond the constraints we have currently considered.

\textbf{Human analysis error rate} The human-in-the-loop approach for LLM discriminators is inherently unscalable due to the manual nature of the analysis and its susceptibility to errors. Therefore, minimizing human involvement by improving LLM discriminators or fostering better collaboration between human experts and LLM discriminators is an important area for future research.

\balance

\section{Conclusion}
In this paper, we propose a novel multi-agent framework that initializes the planning agent by building constraint concepts and enhances its performance through an iterative optimization loop with human and LLM discriminators. Our evaluation demonstrated a 40.2\% improvement in the travel planning problem after just one iteration, highlighting the effectiveness of human discriminators and the human-in-the-loop approach in improving the final pass rate. Given the flexibility of our framework and insights brought about by our preliminary results, we believe it can be extended beyond travel planning to address a broader range of constraint-based problems.
  


\bibliographystyle{ACM-Reference-Format} 
\bibliography{aamas_2024}

\end{document}